\documentclass[runningheads]{llncs}

\usepackage{eccv}

\usepackage{eccvabbrv}

\usepackage{graphicx}
\usepackage{multirow}
\usepackage{booktabs}
\usepackage{xcolor}
\usepackage{tabularx}
\usepackage{float}

\usepackage[accsupp]{axessibility}  

\usepackage{hyperref}

\usepackage{orcidlink}

\begin{document}

\title{``Where am I?''\\ Scene Retrieval with Language } 

\titlerunning{Text2SceneGraph}

\author{Jiaqi Chen\inst{1} \and
Daniel Barath\inst{1} \and
Iro Armeni\inst{2} \and \\
Marc Pollefeys\inst{1,3} \and
Hermann Blum\inst{1,4}}

\authorrunning{J.~Chen et al.}

\newcommand{\customlink}{\href{https://whereami-langloc.github.io}{\texttt{whereami-langloc.github.io}}}

\institute{ETH Zürich \and 
Stanford University \and
Microsoft Mixed Reality \& AI Lab, Zürich \and
University of Bonn
\\
{\scriptsize \customlink}
}

\maketitle

\begin{abstract}
Natural language interfaces to embodied AI are becoming more ubiquitous in our daily lives. 
This opens up further opportunities for language-based interaction with embodied agents, such as a user verbally instructing an agent to execute some task in a specific location. For example, ``put the bowls back in the cupboard next to the fridge'' or ``meet me at the intersection under the red sign.''
As such, we need methods that interface between natural language and map representations of the environment.
To this end, we explore the question of whether we can use an open-set natural language query to identify a scene represented by a 3D scene graph.
We define this task as ``language-based scene-retrieval'' and it is closely related to ``coarse-localization,'' but we are instead searching for a match from a collection of disjoint scenes and not necessarily a large-scale continuous map.
We present Text2SceneGraphMatcher, a ``scene-retrieval'' pipeline that learns joint embeddings between text descriptions and scene graphs to determine if they are a match. The code, trained models, and datasets will be made public.
\end{abstract}

\section{Introduction}
\label{sec:intro}

3D scene graphs have emerged as powerful and efficient representations of spaces~\cite{armeni_sg}, \textit{and they are grounded in language}.
They often describe the semantic information of a scene using natural language labels for the nodes and edges.
Not only can scene graphs efficiently capture the semantics and spatial relations between objects, but they can also efficiently model the hierarchical relationship between spaces, rooms, and buildings. This makes scene graphs well-representative maps of the world.

Furthermore, traditional image-based localization methods often require additional insight in knowing how to capture image queries that are better for the system to localize.
On the other hand, with scene graphs, invoking a text-query is natural and simple for a human--one simply describes their environment.
In this wore bring scene graphs and language into the same paradigm, and propose a method to match scene graphs and open-set language descriptions of the same space.
Trying to localize within scene graphs using language cues also expands existing localization literature to the natural language domain.

Localization with scene graphs and

\begin{figure}[t]
  \centering
  \includegraphics[width=\textwidth]{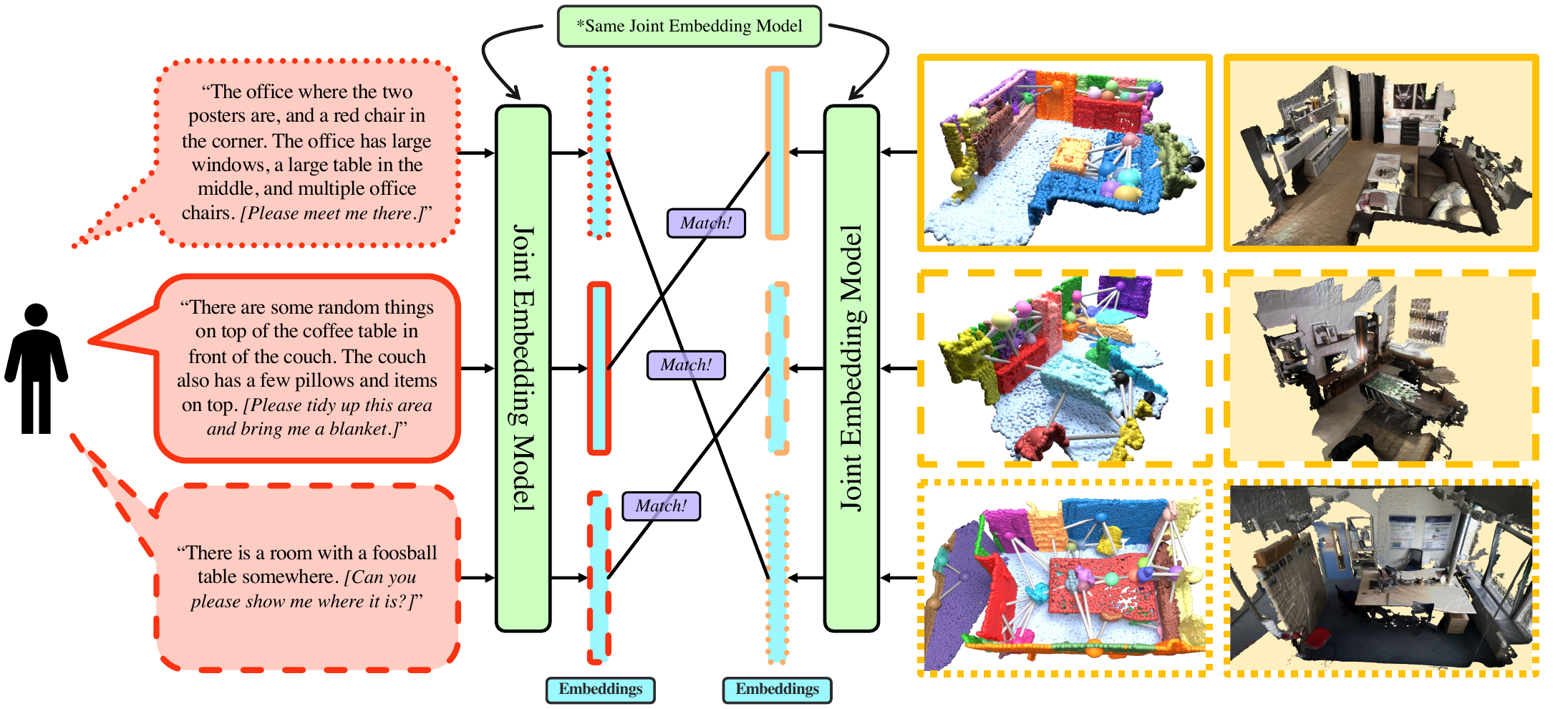}
  \caption{
    \textbf{Pipeline visualization.} Given an open-set natural language query (left, red) and a reference map of environments represented by a set of 3D scene graphs (right, yellow), we establish text-to-scene-graph correspondences.
    The text and scene graph correspondences are matched according to their embeddings in a joint embedding space (blue).
    These embeddings are jointly learned by a joint embedding model (green).
    Additionally, the text-query content in brackets represent potential downstream applications for our system, they are not part of the scene description.
  }
  \label{fig:model_pipeline}
\end{figure}

In particular, this work investigates the task of ``language-based scene-retrieval.''
Given a text of a scene and a set of scene graphs, the task is to identify which scene the text describes.
Additionally, the text-queries are open-set natural language descriptions, meaning they do not follow a specific template or structure.
We consider a scene to be a sub-portion of a larger environment such as an apartment, house, or building.
We also consider scene graphs without metric information; they are ``semantic scene graphs'' where nodes and edges have semantic labels, and represent objects and spatial relationships respectively.
Compared to point clouds or images, scene graphs \textit{naturally} model the object relationships and semantics within the environment.
Tasks similar to this work have recently been proposed for language-prompted retrieval of point clouds~\cite{xia2023text2loc, wang2023text2pos-ret, kolmet2022text2pos-github, scanscribe-3dvista}, or images~\cite{matsuzaki2024cliploc}.
While these works perform ``fine-grained'' localization and output a six degrees-of-freedom (6DoF) pose, ``scene-retrieval'' is more in line with ``coarse-localization,'' which~\cite{xia2023text2loc, wang2023text2pos-ret, kolmet2022text2pos-github} include as an intermediary step in their pipeline.
We propose a method to match open-set text-queries to 3D scene graphs, and compare results to ``coarse-localization'' methods that match based on point clouds or image data of the same scenes.


In our method, dubbed Text2SceneGraphMatcher (Text2SGM), we utilize the close relation between scene graphs and language by first transforming the text-query into a graph of objects and relationships using a large language model (LLM). 
For the remainder of this paper, we will call these ``text-graphs.''
We then train a joint embedding model, to jointly learn latent representations of the scene graphs and text-graphs.
At inference time, we transform the text-query again into a ``text-graph'' and perform retrieval by calculating a matching score against all scene graphs.
The scene and text-query pair with the highest score is then the match.
In this work, we make the following main contributions:
\begin{enumerate}
    \item We introduce a \textit{new} task called \textit{language-based 3D scene retrieval}, which is comparable to coarse-localization using natural language descriptions.  
    \item We propose a method for ``scene retrieval'' which matches open-set natural language text-queries to 3D scene graph representations of scenes.
    \item We further develop a text-to-scene-graph dataset, leveraging scene graphs from 3DSSG \cite{3dssg-wald2020learning}, and pairing these existing scene graphs with new human-generated natural language descriptions of the scenes. We will make this dataset publicly available.
\end{enumerate}

\section{Related Work}
\label{sec:related_work}

\textbf{Localization} 
refers to the task of determining a precise six degrees-of-freedom (DoF) pose of an object or camera within an environment.
Localization methods, such as image-retrieval-based methods~\cite{Germain2019,Germain2020,Sarlin2019,SarlinSuperGlue,Sattler2017,Svarm2017,zeisl2015camera} and pose regression methods~\cite{Kendall2015,Walch2017,Kendall2017,sattler2019understanding,pion2020benchmarking,Moreau2021} require map representations that are either a database of images~\cite{Zhang2006,Zheng2015,pion2020benchmarking,bhayani2021calibrated}, point clouds~\cite{xia2023text2loc, wang2023text2pos-ret, kolmet2022text2pos-github}, or other explicit~\cite{Germain2019,Germain2020,Irschara2009,Li2012,Liu2017,Lynen2020,Sarlin2019,sarlin2021back,Sattler2017,schonberger2018semantic,Svarm2017,zeisl2015camera,SarlinSuperGlue} or implicit 3D representations~\cite{Kendall2015,Valentin2015,Walch2017,Kendall2017, Brachmann2017, Brachmann2018,Balntas2018,Cavallari2019,brachmann2021visual,Moreau2021}.
%
%
Most of these aforementioned works primarily utilize a two-step localization approach where they first \textit{coarsely localize} within the large-scale environment to narrow down the search space and then perform a fine-grained localization step to output the precise 6DoF pose.
We opt for performing ``coarse-localization'' because it does not require geometric information, which our scene-graphs lack by design, as we aim for a lightweight representation.

\noindent
\textbf{Scene Retrieval.} Localizing in a large and dense map representation is typically computationally expensive and memory intensive, which is why some methods adopt a coarse-to-fine procedure~\cite{Sarlin2019, ding2019camnet, ben_image_ret_for_visual_localization,arandjelovic2016netvlad}.
To our knowledge, the closest line of work to coarse-localization is \textit{scene retrieval.}
\cite{3dssg-wald2020learning} proposes the task of image-based 3D \textit{scene retrieval}.
Given a collection of scenes and a single image, they identify the corresponding scene for that image, by calculating a graph edit distance between graph representations of the image and the 3D scene.
In essence, they are coarsely localizing their query image in a map database with disjoint scenes.
We perform a similar task of \textit{language-based 3D scene retrieval}.

\noindent
\textbf{Visual Place Recognition (VPR).}
Another similar line of work is VPR, which aims to simply \textit{identify} a place given some input query~\cite{GargPlaceRecognition, Chen2017_IROS, Garg2019, Hausler2019, Khaliq2020, ji2023cross, gao2023visual, bernreiter2021spherical}.
However, image-based VPR requires high-quality images with sufficient features for matching, while text-queries work better in featureless or low-bandwidth scenarios. 
Images can also pose privacy concerns. 
Also, with language, it is easier to resolve ambiguities in place symmetry.
For example, given a building with symmetric wings, one can simply describe they are in the wing to the right of the main entrance.

\noindent
\textbf{Language \& Localization.}
Recent methods have started to explore the potential use of natural language for identifying one's location in an outdoor point cloud map. 
Text2Pos~\cite{kolmet2022text2pos-github} takes a text description as input, describing the surroundings of the user, and determines the implied position in the point cloud.
Furthermore, Text2Pos utilizes a coarse-to-fine approach. We compare our work against the ``coarse-localizaton'' module in their pipeline.
Text2Loc~\cite{xia2023text2loc} and RET~\cite{wang2023text2pos-ret} improve upon the original Text2Pos method by enhancing the underlying network architecture, \eg, by using a Transformer. 
However, these works do not generalize to indoor scenes. 
Additionally, they do not generalize to an \textit{open-set of human-generated natural language inputs}.

\noindent
\textbf{Scene Graphs.}
Scene representations aim to faithfully model the geometry, semantics, and visual features of a 3D scene, and some focus on semantic representations.
Most notably, Armeni et al.~\cite{armeni_sg}, combine both geometry and semantics into a hierarchical scene graph representation.
Additionally, scene graphs conveniently sit at the intersection of semantics and geometry, combining scene geometry with a more natural language-based representation of the environment.
With the rise of embodied AI, natural language is becoming the modality of choice to interact with these AI agents~\cite{openai2023gpt4, saycan, sayplan, rajvanshi2023saynav, agia2022taskography}.
And already, scene graphs are being used for embodied AI during navigation, task execution~\cite{agia2022taskography, gadre2022continuous, ravichandran2020bridging,jiao2022sequential,Li22a}, and scene manipulation~\cite{dhamo2021graph}.
Particularly in~\cite{sayplan}, scene graphs facilitate a faster semantic search over the entire scene for the objects and areas relevant to the task at hand.
This shows that compared to dense point clouds and meshes, scene graphs can be lightweight while still encoding the general structure of the environment as well as the semantic information.
As we hope to eventually build systems where humans can interact with these embodied AI systems, the communication channels between humans and these agents need to be grounded within a shared environment, setting the stage for semantic scene graphs.

\noindent\textbf{3D Vision-Language Models.} 
Natural language has become a prevalent intermediary between humans and intelligent agents with the rise of Large Language Models (LLMs)~\cite{gpt3, openai2023gpt4} and powerful image-language embedding models~\cite{clip}.
Many 3D Vision-Language Models have emerged for a diverse set of language-grounded tasks such as 3D object identification~\cite{chen2020scanrefer, Achlioptas2020ReferIt3DNL, roh2021languagerefer, guo2023viewrefer, feng2021freeform, huang2022multiview}. 
These works learn mappings between text and objects in a scene~\cite{Achlioptas2020ReferIt3DNL, feng2021freeform, huang2022multiview, guo2023viewrefer, roh2021languagerefer}.
\cite{feng2021freeform} performs node matching between a language graph and a scene graph to find specific object references. 
However, these works operate on a scene level and query for specific objects within a scene.
Other language-grounded tasks include 3D dense captioning~\cite{wang2022spatialityguided, chen2020scan2cap}, visual question and answering~\cite{azuma2022scanqa}, and situated reasoning~\cite{ma2023sqa3d}. 
All in all, these works pave the way for using language to identify scenes.

\section{Methodology}
\label{sec:metho}

\subsection{Problem Formulation}
\noindent\textbf{Scene Retrieval.}
We define ``language-based scene-retrieval'' as the task of finding the correct scene given a natural language description of the scene.
The descriptions can vary in length, but they describe subsets of the objects in a scene, as well as spatial relationships between the objects, such as ``on top of,'' ``next to,'' ``hanging from,'' and ``to the left of.'' 
We propose to tackle the scene retrieval task with a Graph Transformer network inspired by~\cite{transformerConv_shi2021masked}. 
In short, we jointly train the network using a contrastive loss inspired by~\cite{clip} to output an embedding for each text description and each scene graph input.
The output embeddings have a higher cosine similarity if they are from the same scene.
To reinforce the training signal, we also learn a matching probability between 0 and 1, supervised by whether or not the inputs belong to the same scene.

\noindent\textbf{Scene Graph Definition.} 
We define a scene graph $G=(V, E)$ to be a collection of nodes $V$, and edges $E$. 
The nodes are described by feature vectors $\vec{l}\in\mathbb{R}^{m}$, representing the \textit{word2vec} embedding~\cite{word2vec_mikolov2013efficient} of the object label, and any existing attributes.
Each node may also have an associated matrix of object attributes, $\mathbf{A}\in\mathbb{R}^{t \times m}$, such as \textit{color}, \textit{size}, \textit{material}, etc.
More specifically, $\vec{l}=\text{word2vec}(l) + \frac{1}{t}\sum_{i}^{t}\text{word2vec}(\mathbf{A}_{i,:})$, which is simply the sum of the label \textit{word2vec} embedding and the average attribute \textit{word2vec} embedding.
Edges are described by edge feature vectors $\vec{r}\in\mathbb{R}^{m}$, representing the \textit{word2vec} embedding of the the spatial relationship between two objects. 

\begin{figure}[t]
  \centering
  \resizebox{\textwidth}{!}{
    \includegraphics[width=\textwidth]{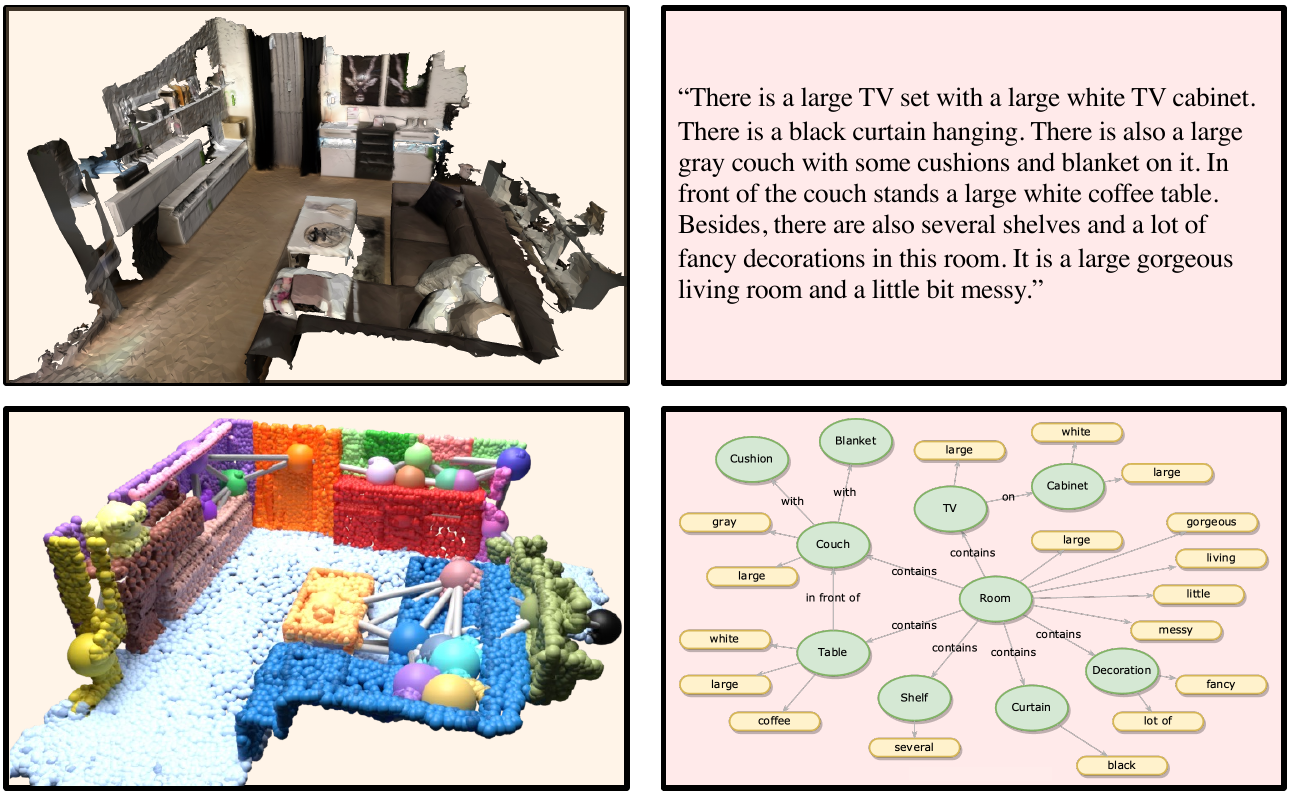}
  }
  \caption{\textbf{Example of Scene and Text Graphs.} In the top left is an image of a living room scene. 
  The bottom left is the corresponding semantic scene graph, where nodes represent the objects, and edges represent a spatial relationship such as ``contains,'' ``in front of,'' or ``on.'' 
  The top right section shows a text description of the scene, from the human-annotated dataset. 
  The bottom right figure shows the corresponding ``text-graph'' of the scene description. 
  The left hand side represents our database of scene graphs, while the right hand side represents an incoming text-query, which we first transform into a ``text-graph,'' and then match with the scene graph.}
  \label{fig:json_graphs}
\end{figure}

We specifically do not include 3D position, point clouds, or metric information in our scene graph to minimize memory footprint and improve potential scalability.
A visual example of this semantic scene graph can be seen in Figure~\ref{fig:json_graphs}. More examples can be found in the supp.\ m.

\subsection{Data Processing}
We first transform the text-queries into graph representations by extracting the objects and relationships -- we call these ``text-graphs.''
We also process the scene graphs to only include edges within some distance threshold.
Then, for both the text-graphs and scene graphs, we vectorize the node and edge labels into fixed-length feature vectors.

\noindent\textbf{ScanScribe \& Human Dataset to Graphs.}
We first transform the text-queries into a ``text-graph.'' 
To create a text-graph from a description, we use a large language model (LLM), GPT-4~\cite{openai2023gpt4}, to extract the objects, attributes, and relationships.
A description may say, ``There is a wooden chair next to a table,'' in which case the extracted graph would have nodes representing ``chair'' and ``table,'' and an edge in between representing the relationship, ``next to.''
Edges are only defined if a relationship is stated in the description.
We prompt the LLM to extract the relevant information from the text descriptions, and output only JSON format.
Examples of the prompt, inputs, and outputs are in the supp.\ m. 
A visual example of a ``text-graph'' can be seen in Figure~\ref{fig:json_graphs}. 

\noindent\textbf{3DSSG Graphs.}
The 3D scene graphs in our datasets are densely connected, therefore we process our scene graphs to only include edges that are within some distance threshold, set to 1.5m. This allows us to only consider relationships between neighboring objects.
The distance between two objects is calculated as the nearest distance between the two object bounding boxes.

\noindent\textbf{Node/Edge Label Vectorization.}
The last step involves transforming the node and edge labels into feature vectors.
For this we use \textit{word2vec}~\cite{word2vec_mikolov2013efficient} to simply encode each label, which is a string, into a feature vector of length 300.

\subsection{Model Architecture}

\begin{figure}[t]
  \centering
  \includegraphics[width=\textwidth]{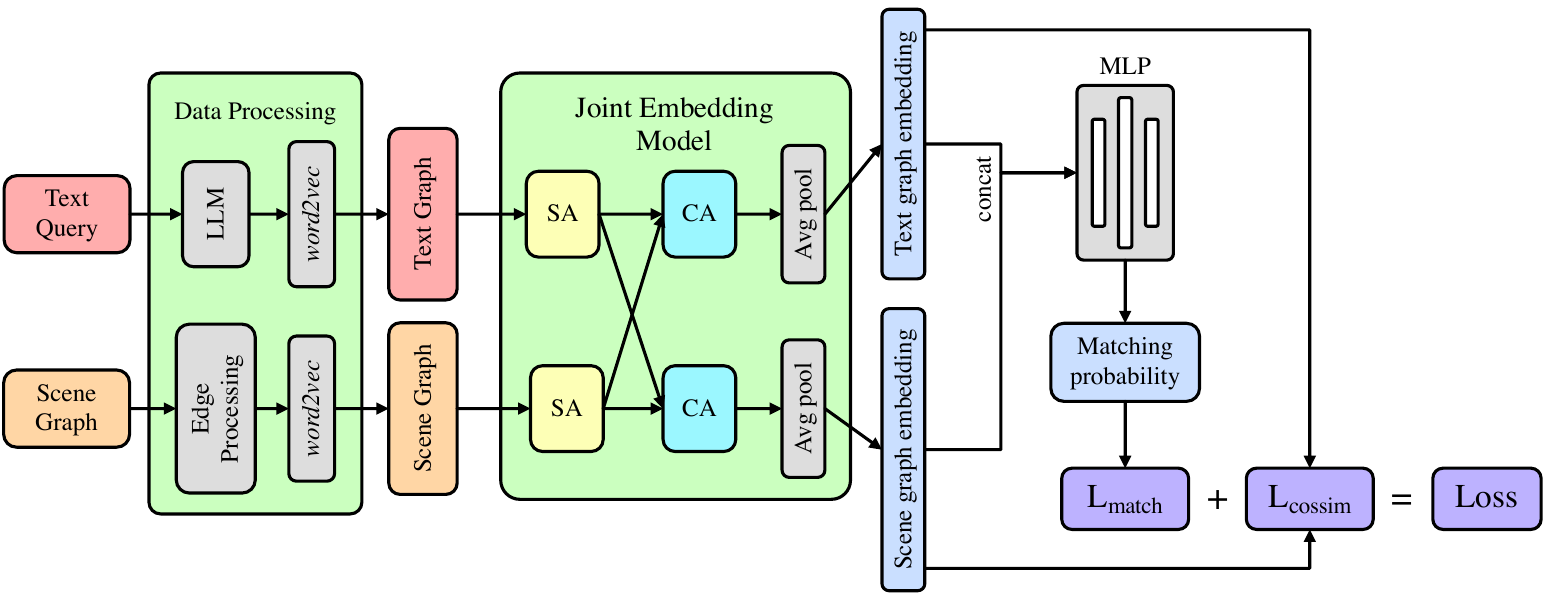}
  \caption{\textbf{Pipeline Overview.} The input to our method is a text-query and a 3D scene graph potentially matching the text. Next, we independently process these inputs to obtain graphs with \textit{word2vec} embeddings as nodes for the objects. Our network then performs self- and cross-attention with a final average pooling layer to obtain embeddings. 
  These embeddings are concatenated, and a matching probability is predicted by a Multi-Layer Perceptron (MLP).   
  The joint embedding model is trained by jointly optimizing a matching loss and cosine similarity loss. 
  }
  \label{fig:model_pipeline}
\end{figure}

An overview of the pipeline is in Figure~\ref{fig:model_pipeline}.
After transforming our inputs into a standardized graph format, we pass them into the joint embedding model.
The first step is an attention module with a self-attention and then a cross-attention layer.
The self-attention and cross-attention layers are implemented with the graph transformer operator inspired by~\cite{transformerConv_shi2021masked}.
In the self-attention module, the message passing happens between edges that exist within the text- or scene-graph. 
In the cross-attention module, if we have a text-graph with nodes A and B connected with an edge, and a scene-graph with C and D, then during cross-attention, message passing would happen between nodes (A, C), (A, D), (B, C), (B, D). 
After messaging passing, the nodes are aggregated by addition.
Finally, the graph nodes are average pooled across all node features, into a final graph embedding vector for each text-graph and scene graph, $\vec{S}_{text}$ and $\vec{S}_{scene}$. 
Then, $\vec{S}_{text}$ and $\vec{S}_{scene}$ are concatenated into one long vector before being passed into a 3-layer MLP that outputs a matching probability between the two embeddings.
The overall loss is a combination of a ``matching probability'' loss and a ``cosine similarity'' loss.
The same architecture is used during training and inference.

We choose the number of self-attention/cross-attention modules $N$ to be 1 for our purposes because too many propagation iterations lead to over-smoothing and the graph converges to the same feature vector everywhere--a known problem in graph neural networks~\cite{oversmoothing-chen2019measuring, oversmoothing-rusch2023survey}, especially detrimental when graph size is small.
Note that due to the cross attention module, the model requires a text-query \textit{and} a scene graph as input.
But as we show in Sec.~\ref{sec:results}, regardless of the input pair, the model outputs a sufficiently meaningful embedding.

\subsection{Contrastive Loss}
Our loss is the sum of the ``matching probability'' and ``cosine similarity'' terms.
The cosine similarity loss term is calculated as follows:

%
\begin{equation}
    L_{\text{cossim}} = \frac{1}{B^2} \sum_{i} \sum_{k} -w_{ik} \log \left( \frac{\exp(x_{ik})}{\sum_{j} \exp(x_{ij})} \right),
    \label{eq:loss_cos_sim}
\end{equation}
where
\begin{equation}
    x = 1 - \text{cos\_sim}(\vec{S}_{\text{scene}}, \vec{S}_{\text{text}}),
    \label{eq:loss_cos_sim_x}
\end{equation}
and $w_{ik}$ is the target $(1 - \text{cosine similarity})$ between the $i^{th}$ scene graph embedding vector $\vec{S}_{scene}$, and $k^{th}$ text-graph embedding vector $\vec{S}_{text}$.
The matching probability loss term is formulated as follows:
\begin{equation}
    L_{\text{match}} = \frac{1}{B^2} \sum_{i} \sum_{k} -y_{ik} \log \left( \frac{\exp(m_{ik})}{\sum_{j} \exp(m_{ij})} \right),  \quad m \in [0, 1],
    \label{eq:loss_match}
\end{equation}
where $m$ is the output matching probability, and $y_{ik}$ is the target matching probability between the $i^{th}$ scene graph and the $k^{th}$ text-graph. 
$B$ is the batch size, and $j$ indexes through the text-graphs. 
Our overall loss is the following:
\begin{equation}
    L = \frac{1}{2}(L_{\text{match}} + L_{\text{cossim}}).
\end{equation}
This allows us the quantify how similar are a text and scene graph.

\begin{table}[H]
  \caption{
  The recall for matching the top $k$ out of 10 scene candidates to our text-query. 
  These candidates are sampled from the ScanScribe (top)~\cite{scanscribe-3dvista} and the collected Human (bottom) datasets. 
  Together with the baselines, Text2Pos and CLIP2CLIP, we show three versions of our pipeline, ``match-prob'' and ``cos-sim'' employing different \textbf{matching score calculations}, and ``ret-based'' (precomputing scene graph embeddings and doing retrieval-based matching). Note that the underlying joint embedding models in all three models are trained with the matching probability \textit{and} cosine similarity in the loss function. They differ in how they calculate the \textbf{matching score}.
  }
  \label{tab:recall}
  \centering
  \scriptsize
    \begin{tabular}{@{}lcccc@{}}
      \toprule
      Top k  & 1 & 2 & 3 & 5  \\
      \midrule
      & \multicolumn{4}{c}{\textbf{ScanScribe}}   \\ 
      Text2Pos (from scratch)       & $11.63\pm0.99$  & $22.57\pm1.16$  & $33.03\pm1.36$  & $52.80\pm1.67$  \\
      Text2Pos (fine-tuning)        & $10.30\pm0.93$  & $20.74\pm1.32$  & $31.65\pm1.45$  & $53.42\pm1.68$  \\
      CLIP2CLIP                     & $33.09\pm0.21$  & $52.53\pm0.24$  & $65.98\pm0.21$  & $82.69\pm0.13$  \\
      \textbf{Text2SGM} (\textit{match-prob}) & $63.70\pm1.85$  & $81.40\pm1.85$  & $89.31\pm1.42$  & $95.71\pm0.93$  \\
      \textbf{Text2SGM} (\textit{cos-sim})    & $68.27\pm2.05$  & $84.95\pm1.62$  & $\mathbf{91.65\pm1.23}$  & $\mathbf{97.14\pm0.76}$  \\
      \textbf{Text2SGM} (\textit{ret-based})  & $\mathbf{68.61\pm0.04}$  & $\mathbf{85.00\pm0.02}$  & $91.57\pm0.01$  & $96.99\pm0.00$  \\
      \midrule
      & \multicolumn{4}{c}{\textbf{Human}}        \\ 
      Text2Pos (from scratch)       & \phantom{0}$9.57\pm0.95$   & $20.82\pm1.27$  & $32.10\pm1.44$  & $52.93\pm1.48$  \\
      Text2Pos (fine-tuning)        & \phantom{0}$7.38\pm0.75$   & $15.41\pm1.21$  & $24.82\pm1.37$  & $45.93\pm1.60$  \\
      CLIP2CLIP                     & $49.87\pm0.16$  & $71.09\pm0.13$  & $83.07\pm0.09$  & $\mathbf{95.11\pm0.03}$  \\
      \textbf{Text2SGM} (\textit{match-prob}) & $51.04\pm2.27$  & $70.86\pm2.20$  & $81.50\pm1.79$  & $91.48\pm1.23$  \\
      \textbf{Text2SGM} (\textit{cos-sim})    & $\mathbf{53.90\pm2.37}$  & $\mathbf{74.98\pm2.28}$  & $\mathbf{85.85\pm1.54}$  & $94.87\pm0.94$  \\
      \textbf{Text2SGM} (\textit{ret-based})  & $53.45\pm0.05$  & $74.52\pm0.04$  & $85.44\pm0.03$  & $94.60\pm0.01$  \\
      \bottomrule

    \end{tabular}
\end{table}

\section{Experiments}
\label{sec:exp}
\subsection{Datasets}

\noindent
\textbf{ScanScribe Dataset.} 
The ScanScribe dataset~\cite{scanscribe-3dvista} is the first scene-text pair dataset generated for pre-training a 3D vision-language grounding model.
They source the underlying scenes from the ScanNet~\cite{dai2017scannet} and 3RScan~\cite{3rscan-rio} dataset, and leverage the pre-existing text annotations from ScanQA~\cite{azuma2022scanqa}, ScanRefer~\cite{chen2020scanrefer}, and ReferIt3D~\cite{Achlioptas2020ReferIt3DNL}, as well as scene graph annotations from 3DSSG~\cite{3dssg-wald2020learning} to generate scene descriptions.
The authors of the ScanScribe dataset generate these scene descriptions by prompting GPT-3~\cite{gpt3}, using templates, and compiling the existing text annotations.

We only use a subset of this ScanScribe dataset, one that includes both scene descriptions and scene graph representations from 3DSSG~\cite{3dssg-wald2020learning}.
For the remainder of the paper, when we refer to the ``ScanScribe dataset,'' we refer to this subset of the larger ScanScribe dataset~\cite{scanscribe-3dvista}.

For each scene in the ScanScribe dataset, there are on average 20 text descriptions, leading to a total of 4472 descriptions that describe 218 unique scenes.
Descriptions in this ScanScribe dataset are generated using GPT-3~\cite{gpt3}.
The text descriptions also do not always describe the scene to its entirety, as the authors first sample varying sizes of subgraphs from the scene graphs, and then generate a description.
Examples of these text descriptions and their corresponding text graphs can be found in supp.\ m.
We separate this ScanScribe dataset into a training set and a testing set, where the testing set has 1116 descriptions from 55 unique scenes.
This split will be made public for full reproducibility.

\noindent
\textbf{Human-Annotated Dataset.} \textcolor{red}{
}
We ask human annotators to observe sequences of images from a scene within the 3RScan dataset~\cite{3rscan-rio}, and then write a short description of what they see, focusing on objects and spatial relationships.
These scenes already have corresponding scene graphs from the 3DSSG~\cite{3dssg-wald2020learning} dataset.
The human annotators are asked to describe the scene in a natural, ``human-like'' way.
We collect 147 natural language descriptions of 142 unique scenes in this way.
We consider this human-annotated dataset as another testing set and refer to it from now on as the ``Human dataset.''
These descriptions were \textit{not} generated from the scene-graph, thus there is no bias.
Examples of these text descriptions and their corresponding text graphs can be found in Figure~\ref{fig:json_graphs}.

\begin{table}[t]
  \caption{The recall values for matching the top $k$ scenes, \textbf{out of all scenes}, to our text-queries. All scene candidates are from the ScanScribe (top)~\cite{scanscribe-3dvista} and the collected Human (bottom) datasets.
  Together with the baselines, Text2Pos and CLIP2CLIP, we show three versions of our pipeline, ``match-prob'' and ``cos-sim'' employing different \textbf{matching score calculations}, and ``ret-based'' (precomputing scene graph embeddings and doing retrieval-based matching). Note that the underlying joint embedding models in all three models are trained with the matching probability \textit{and} cosine similarity in the loss function. They differ in how they calculate the \textbf{matching score}.
}
  \label{tab:recall_all_scenes}
  \centering
  \scriptsize
    \begin{tabular}{@{}lcccc@{}}
      \toprule
      Top k  & 5 & 10 & 20 & 30  \\
      \midrule
      & \multicolumn{4}{c}{\textbf{ScanScribe, 55 scenes}}   \\     
      Text2Pos (from scratch)        
                                     & $10.27\pm0.83$    & $21.24\pm1.17$    & $39.36\pm1.43$   & $57.30\pm1.48$    \\
      Text2Pos (fine-tuning)         
                                     & \phantom{0}$9.21\pm0.92$     & $18.27\pm1.24$    & $39.27\pm1.61$   & $58.63\pm1.60$    \\
      CLIP2CLIP                      
                                     & $33.26\pm0.21$    & $53.47\pm0.23$    & $73.76\pm0.18$   & $86.23\pm0.11$    \\
      \textbf{Text2SGM} (\textit{match-prob}) & $70.24\pm1.93$    & $85.71\pm1.38$    & $93.36\pm1.06$   & $97.25\pm0.65$    \\
      \textbf{Text2SGM} (\textit{cos-sim})    & $\mathbf{76.34\pm2.06}$    & $\mathbf{87.83\pm1.55}$    & $\mathbf{95.40\pm0.89}$   & $\mathbf{98.24\pm0.57}$    \\
      \textbf{Text2SGM} (\textit{ret-based})  & $76.29\pm0.16$  & $87.77\pm0.10$  & $95.34\pm0.05$  & $98.18\pm0.02$   \\
      \midrule
      Top k  & 5 & 10 & 30 & 75  \\
      \midrule
      & \multicolumn{4}{c}{\textbf{Human, 142 scenes}}        \\ 
      Text2Pos (from scratch)        & $2.03\pm0.47$   & $6.68\pm0.85$   & $21.69\pm1.32$  & $54.30\pm1.59$  \\
      Text2Pos (fine-tuning)         & $2.76\pm0.45$   & $5.40\pm0.70$   & $13.51\pm1.09$  & $48.99\pm1.41$  \\
      CLIP2CLIP                      & $36.18\pm0.08$  & $47.56\pm0.08$  & $75.54\pm0.06$  & $\mathbf{98.64\pm0.00}$   \\
      \textbf{Text2SGM} (\textit{match-prob}) & $\mathbf{38.91\pm2.86}$  & $51.31\pm2.74$  & $78.16\pm1.81$  & $91.34\pm0.39$   \\ 
      \textbf{Text2SGM} (\textit{cos-sim})    & $34.81\pm2.59$  & $55.84\pm2.52$  & $83.03\pm4.71$  & $95.87\pm0.88$
   \\
      \textbf{Text2SGM} (\textit{ret-based})  & $35.07\pm0.20$  & $\mathbf{56.85\pm0.28}$  & $\mathbf{84.04\pm0.14}$  & $95.88\pm0.04$   \\
      \bottomrule
    \end{tabular}
\end{table}

\subsection{Evaluation Setup}

For all mentioned models, we first train on a training set. We also have a validation set that we use to determine the best model parameters. We then retrain our model on the combined training and validation set to maximize the amount of training data. Finally, we test on completely separate testing sets.

\noindent\textbf{Text2Pos.} 
We compare our method to the ``coarse-localization'' module from Text2Pos~\cite{kolmet2022text2pos-github}.
This module performs coarse-localization in a large-scale outdoor point cloud using a natural language query.
Text2Pos does this by separating a large scale point cloud into ``cells'' and identifies the top $k$ most likely ``cells.''
The ``cells'' and text-query are first individually embedded by a point cloud embedding model and a text embedding model respectively, and then matched according to their cosine similarity.
Since Text2Pos takes point cloud inputs, we extract the point clouds from the corresponding scenes in our datasets and consider each scene as a single ``cell.''
Notably, other concurrent work, Text2Loc~\cite{xia2023text2loc} and RET~\cite{wang2023text2pos-ret} build off of Text2Pos, but their models are not available.
We compare against a fine-tuned and retrained version of the ``coarse-localization'' module from Text2Pos, trained on a training set from ScanScribe dataset.

In order to encode the text-queries, Text2Pos trains a LanguageEncoder.
This LanguageEncoder requires, as input, a predefined mapping for all ``known-words'' in the text-queries of the dataset to a unique number.
Text2Pos first vectorizes the text-queries with this mapping, and then feeds it into an LSTM for the embedding.
It is worth noting that the ``coarse-localization'' module needs to be trained with this mapping.
This is a disadvantage compared to our method in that we do not require such a mapping.
The Text2Pos method does not generalize well to text-queries that are not yet previously mapped.

\begin{table}[t]
  \caption{\textbf{Runtime and Memory.} We show the evaluation runtime and memory storage costs for different methods. 
  The runtime is represented as time in seconds. 
  The memory storage is represented as MB. 
  We calculate runtime as the time it takes for one text-query to be embedded and matched against 10 pre-processed scene graph embeddings to find a best match. 
  And we calculate memory storage for the entire database of scene graph embeddings (ScanScribe and Human Datasets).
  This simulates the most efficient storage scenario where we store a reference map of embeddings, computed offline, and run nearest-neighbors search for matches.}
  \label{tab:timing}
  \centering
  \scriptsize
  \begin{tabular}{@{}lcccc@{}}
    \toprule
      & \multicolumn{2}{c}{Runtime (s)} & \multicolumn{2}{c}{Memory (MB)} \\
      \cmidrule(lr){2-3} \cmidrule(lr){4-5} 
      Model  & ScanScribe & Human & ScanScribe & Human  \\
    \midrule
    Text2Pos   & 0.00918 & 0.01652           & 0.11 & \phantom{0}0.29  \\
    CLIP2CLIP  & 0.03107 & 0.05623           & 5.63 & 14.54 \\
    Text2SGM    & 0.03639 & 0.01206          & 0.13 & \phantom{0}0.34  \\
  \bottomrule
  \end{tabular}
\end{table}

\noindent\textbf{CLIP2CLIP.}
A second baseline we compare to is one that we implement ourselves, and we call it CLIP2CLIP.
The idea is as follows: we embed the text-queries and images from the scenes using CLIP~\cite{clip}, and then attempt to match text description to scenes via these CLIP embeddings.
The context size of CLIP's text encoder only supports lengths up to 77, which is why we first separate the sentences into phrases delimited by commas or periods.
Then we encode each smaller phrase, and take the average over all the phrases as the ``text-query encoding.''
For the images, we simply encode each image using CLIP's image encoder.
Then, in order to get a matching score between a ``text-query'' and a scene of images, we calculate the cosine similarity between the text encoding and all image encodings from the scene, and take the maximum score as the matching score between that text and scene.
During inference, given a ``text-query,'' we encode it and calculate the cosine similarity between the text encoding and image encoding of each scene.
The highest scoring pair is the match.
The CLIP model is not fine-tuned or retrained.

\noindent\textbf{Text2SGM.}
We present two procedures for calculating the \textbf{\textit{matching score}}.
The first (match-prob) uses the ``matching probability'' outputs from the model as the matching scores between a text-query and a scene graph.
The second one (cos-sim) calculates a cosine similarity between the text-query and scene graph embedding produced by the model.
Both procedures use the same underlying joint embedding model.

We also present a third retrieval-based procedure for finding text and scene graph matches.
Given that our pipeline requires an input \textit{pair} of a text-query and a scene graph--due to the cross attention module--we investigate how different text-queries affect the resulting scene graph embeddings and vice versa. 
Detailed results of this investigation are in the supplementary material, but we observe that the embeddings are still consistently identifiable, despite being paired with different inputs.
As such, for the third version of our model, we fix the opposing paired input using the same text-query or scene graph.
This text-query or scene graph is randomly sampled from the training dataset, and we embed all inputs with this fixed counterpart.
Given a text-query, we pass it to the model along with a fixed scene graph input, and get the text-graph embedding.
Then we find the highest cosine similarity match in a database of pre-computed scene graph embeddings.
We call this our retrieval-based (ret-based) matching approach.

To reiterate, the underlying joint embedding model is the same across all three ``matching score'' calculation procedures.
The model is trained with \textbf{both} the matching probability and cosine similarity loss as a part of the objective function.
The ``match-prob'' procedure uses the matching probability as the matching score.
The ``cos-sim'' procedure uses the cosine similarity between embeddings as the matching score.
The ``ret-based'' procedure also uses the cosine similarity as the matching score, but pre-computes a database of scene graph embeddings.

\subsection{Metrics}

We measure two recall metrics, the first measures a ``top k out of 10'' recall.
We try to match one randomly sampled text-query against 10 scene graphs, out of which 1 scene graph is the correct one.
We measure if the correct pair is matched within the top 1, 2, 3, and 5 matches out of 10.
Similarly, we measure a ``top k out of the entire dataset'' recall, where we try to retrieve the correct graph from the entire dataset.
We report recall values in Table~\ref{tab:recall} and Table~\ref{tab:recall_all_scenes}.
In addition to recall, we also measure the inference time of matching 1 text-query to a scene graph.
We report these times in Table~\ref{tab:timing}.
Lastly, for each method, we calculate the size of storing our datasets in memory.
More specifically, we calculate the embedding size of each method, since these scene embeddings can be precomputed (ret-based method) and stored for later retrieval.

\subsection{Results}
\label{sec:results}

We first present the recall values for the three versions of our pipeline compared against baselines on the ScanScribe and the captured Human datasets~\cite{scanscribe-3dvista}.

Table~\ref{tab:recall} reports the recalls when performing top-$k$ selection from $10$ candidates scenes. 
It also shows that our method outperforms Text2Pos as well as CLIP2CLIP by a large margin on both datasets in almost all configurations.
Only in the Human dataset, for the top 5 out of 10 and the top 75 out of the entire dataset (Table~\ref{tab:recall_all_scenes}), does CLIP2CLIP outperform our method.
For the most part, using the embedding cosine similarity as the scoring mechanism performs the best.
We posit that this is due to the high-dimensional joint embedding capturing the nuances in the graphs, something not achieved by using just the matching probability (\ie, a single real number).
We additionally recognize that the requirement for a ``known-words'' mapping in Text2Pos severely limits its ability to generalize to the open-set text-queries from the Human dataset.
Furthermore, these results show that Text2Pos does not generalize to indoor scenes.

Our retrieval-based pipeline performs almost as well as, and in some instances better than, the pipeline (cos-sim) that jointly calculates the embeddings.
This suggests our joint embedding model learns an adequately meaningful embedding.
Also, we observe that the CLIP2CLIP performance is much more competitive to our performance on the Human dataset, compared to the ScanScribe dataset.
Seeing as our training data is from the same distribution as the ScanScribe dataset, our much better performance here is to be expected.
However, given that CLIP~\cite{clip} is pre-trained on a much larger-scale dataset, we also see that it generalizes better on the open-set text-queries from the Human dataset.
Nevertheless, these recall values suggest that our method successfully retrieves scenes given a text-query describing the scene.


Results when performing top-$k$ selection from \textit{all} candidates scenes are reported in Table~\ref{tab:recall_all_scenes}. 
Text2Pos performs on the level of random selection for both scenes in all recall values.
The proposed Text2SGM method improves upon this significantly, achieving meaningful results across almost all datasets and configurations.
The highest recall is achieved, again, by employing the cosine similarity as the scoring mechanism.
The retrieval-based approach does not lag much behind in terms of recall, suggesting that the obtained graph embeddings are meaningful despite using the same input text-query, and thus, generalize well. 
Additionally, we show results in Table~\ref{tab:gen_from_gpt} using text-queries that are generated by GPT-4~\cite{openai2023gpt4} from images of the 3DSSG scenes.
Even though we are now biasing towards the CLIP2CLIP method since we generate our text-queries from images, we still perform more accurately.

\begin{table}[t]
    \caption{The recall values from using text-queries generated by GPT-4 from images of the 3DSSG scenes. 
            We generate text-queries from images because the ScanScribe test set in Tables~\ref{tab:recall} and~\ref{tab:recall_all_scenes} has queries generated from the scene graphs.
            The Human dataset, in addition to this experiment, eliminates that bias.
            We measure top-k out of 10 recall on the ScanScribe test set, note that we generate 1 text-query per scene.}
    \centering
    \scriptsize
        \begin{tabular}{@{}lcccc@{}}
          \toprule
          Top k  & 1 & 2 & 3 & 5  \\
          \midrule
          CLIP2CLIP                                 & $33.07\pm0.21$  & $52.52\pm0.23$  & $66.00\pm0.21$  & $\textbf{82.78}\pm0.13$   \\
          \textbf{Text2SGM} (\textit{cos-sim})      & $\textbf{34.22}\pm1.77$  & $\textbf{56.67}\pm1.70$  & $\textbf{68.78}\pm2.35$  & $82.11\pm1.23$  \\
          \bottomrule
        
        \end{tabular}
    \label{tab:gen_from_gpt}
\end{table}

\subsection{Runtime and Storage Requirements}

The average runtime and storage requirements of the tested algorithms on both datasets are reported in Table~\ref{tab:timing}.
All tested methods run in real time and approximately at similar speeds, \ie, in a few tens of milliseconds at most.
The reference map representation only takes a few hundred kilobytes for the proposed method and Text2Pos, while CLIP2CLIP requires at least an order of magnitude more storage.
These results underscore the practicality of the proposed approach in bandwidth-limited scenarios requiring real-time performance.

\subsection{Ablation Studies}

In this section, we perform several ablation studies in order to gain a more nuanced understanding of the proposed method. 

Table~\ref{tab:loss_ablation} reports the recall values together with their standard deviations on both tested datasets, employing cosine similarity ($\theta$), matching probability ($m$) or both ($\theta, m$),\textbf{ in the loss objective}, when \textit{training} the proposed network.
Interestingly, while training with only the matching probability lags behind other losses on the ScanScribe dataset, it leads to the best performance on the Human dataset.
Conversely, a cosine similarity loss alone leads to the most accurate results on ScanScribe, while being marginally less accurate on the Human dataset.
Employing both losses leads to the best results by a large margin.

Table~\ref{tab:saca_ablation} reports the results when varying the number of self and cross-attention modules in the proposed method. 
The best results are achieved with only one module, with every additional one significantly reducing the recall. 
This could be due to over-smoothing problem~\cite{oversmoothing-chen2019measuring, oversmoothing-rusch2023survey} as previously mentioned.
Thus, we use a single module in all experiments. 


\begin{table}[t]
  \caption{\textbf{Loss Ablation.} Since our model outputs embedding vectors as well as a matching probability, we run an ablation on the loss objective function. More specifically, $\theta$ represents using only the cosine similarity of the embedding vectors. $m$ represents using only the matching probability as the loss. And $\theta$, $m$ represents using an average of both the cosine similarity and matching probability as the loss. We show recall values for the ScanScribe and Human datasets, at a top $k$ out of 10 precision.}
  \label{tab:loss_ablation}
  \centering
  \scriptsize
  \resizebox{\textwidth}{!}{
  \begin{tabular}{@{}lcccccccc@{}}
    \toprule
    \multirow{2}{*}{Top k} & \multicolumn{4}{c}{ScanScribe} & \multicolumn{4}{c}{Human} \\
    \cmidrule(lr){2-5} \cmidrule(lr){6-9}
    out of 10 & 1 & 2 & 3 & 5 & 1 & 2 & 3 & 5 \\
    \midrule
    & \multicolumn{4}{c}{\textbf{ScanScribe}} \\
    $\theta$        & $65.17\pm4.83$   & $79.99\pm4.24$  & $86.35\pm3.90$  & $93.53\pm2.69$  & $44.88\pm4.97$  & $64.63\pm4.82$  & $76.12\pm4.02$  & $89.52\pm2.91$ \\
    $m$             & $54.16\pm5.56$   & $74.47\pm4.59$  & $84.34\pm3.61$  & $93.36\pm2.53$  & $47.92\pm4.67$  & $68.54\pm4.67$  & $79.60\pm4.31$  & $91.52\pm2.72$ \\
    $\theta$, $m$   & $62.70\pm4.71$   & $80.88\pm3.81$  & $88.07\pm3.22$  & $95.16\pm2.15$  & $51.38\pm4.98$  & $73.79\pm4.55$  & $84.16\pm3.60$  & $92.02\pm2.64$ \\

  \bottomrule
  \end{tabular}
  }
\end{table}

\begin{table}[t]
  \caption{\textbf{Number $N$ of Self-Attention/Cross-Attention Module Ablation.} We show recall values for adding multiple layers of the self-attention/cross-attention module during training. These are recall values for the ScanScribe and Human datasets, at a top $k$ out of 10 precision.}
  \label{tab:saca_ablation}
  \centering
  \scriptsize
  \resizebox{\textwidth}{!}{
  \begin{tabular}{@{}lcccccccc@{}}
    \toprule
    \multirow{2}{*}{Top k} & \multicolumn{4}{c}{ScanScribe} & \multicolumn{4}{c}{Human} \\
    \cmidrule(lr){2-5} \cmidrule(lr){6-9}
    out of 10 & 1 & 2 & 3 & 5 & 1 & 2 & 3 & 5 \\
    \midrule
    N=1  & $62.70\pm4.71$ & $80.88\pm3.81$ & $88.07\pm3.22$ & $95.16\pm2.15$ & $51.38\pm4.98$ & $73.79\pm4.55$ & $84.16\pm3.60$ & $92.02\pm2.64$ \\
    N=2  & $56.16\pm5.12$ & $76.58\pm4.14$ & $84.90\pm3.88$ & $92.96\pm2.87$ & $45.05\pm4.76$ & $66.13\pm4.03$ & $77.40\pm3.93$ & $88.01\pm2.82$ \\
    N=4  & $23.02\pm4.30$ & $44.16\pm4.71$ & $59.88\pm5.16$ & $78.76\pm3.80$ & $16.31\pm3.71$ & $31.80\pm5.10$ & $45.46\pm5.37$ & $66.52\pm5.47$ \\
  \bottomrule
  \end{tabular}
  }
\end{table}


\section{Conclusion}
\label{sec:conclu}
In conclusion, we present Text2SceneGraphMatcher for language-based scene retrieval. 
We demonstrate that open-set natural language text-queries are able to retrieve the corresponding scene they describe. 
We do so by training a joint embedding model on the text-queries and scene graph representations of the scenes, and find matches using the embeddings.
We also demonstrate our method's generalizability on open-set human-annotated text-queries.
We present a human-annotated dataset of text-to-scene-graph pairs, and show that we are still able to retrieve scene and text matches.
Future work include extending the scene graphs to larger-scale hierarchical environments and performing fine-grained localization.
The code, trained models, and datasets will be made public.

\noindent
\textbf{Acknowledgements}. This work was partially funded by the ETH RobotX research grant, the Hasler Stiftung Research Grant via the ETH Zurich Foundation, SBB (ETH Mobility Initiative Partner).

%
%
\bibliographystyle{splncs04}
\bibliography{egbib}
\end{document}